# Dual Control Memory Augmented Neural Networks for Treatment Recommendations

Hung Le, Truyen Tran and Svetha Venkatesh

Applied AI Institute, Deakin University, Geelong, Australia
`{lethai,truyen.tran,svetha.venkatesh}@deakin.edu.au`

**Abstract.** Machine-assisted treatment recommendations hold a promise to reduce physician time and decision errors. We formulate the task as a sequence-to-sequence prediction model that takes the entire time–ordered medical history as input, and predicts a sequence of future clinical procedures and medications. It is built on the premise that an effective treatment plan may have long–term dependencies from previous medical history. We approach the problem by using a memory–augmented neural network, in particular, by leveraging the recent differentiable neural computer that consists of a neural controller and an external memory module. But differing from the original model, we use *dual controllers*, one for encoding the history followed by another for decoding the treatment sequences. In the encoding phase, the memory is updated as new input is read; at the end of this phase, the memory holds not only the medical history but also the information about the current illness. During the decoding phase, the memory is *write–protected*. The decoding controller generates a treatment sequence, one treatment option at a time. The resulting dual controller write–protected memory–augmented neural network is demonstrated on the MIMIC-III dataset on two tasks: procedure prediction and medication prescription. The results show improved performance over both traditional bag-of-words and sequence-to-sequence methods.

## 1 Introduction

A core task in healthcare is to generate effective treatment plans. Machine-assisted treatment recommendations thus have potential to improve healthcare efficiency. We approach the task by learning from rich electronic medical records. An electronic medical record (EMR) is a digital record of patient health information over time such as details of symptoms, data from monitoring devices, and clinicians' observations. Among these data elements, diagnosis, clinical procedure and drug prescription codes are core information, and they are highly temporal correlated. A medical history is a sequence of clinic visits, each of which, at minimum, has a set of diagnoses, a set of treatment procedures, and a set of discharge medications. In MIMIC-III dataset [9], diagnoses are "ordered by priority", procedures follow the order that "the procedures were performed" and the drugs follow the date of prescriptions[1]. The temporal dependency in EMR clinical codes can be long-term. For example, once diagnosed with diabetes (Type I or II), the conditions (and hence its medications, if any) are persistent through the patient's life, even though it might not be coded at every visit. Since EMR data are temporally sequenced by patient medical visits, clinical codes at current admission may be related to other codes appearing in previous admissions.

These long-term dependencies pose a great challenge for prediction models. Recent efforts dealing with medical prediction have largely focused on modeling the admission's diagnoses and treatments as two sets of codes and capture sequential dependencies between codes from different admissions, i.e., sequence of sets [3,11,13,14,15]. This approach may expose limitations since using admission set representation ignores the internal sequential dependencies and thus fails to discover sequential relations among codes from the same admission.

To tackle these issues, we propose a novel treatment recommendation model using a memory-augmented neural network (MANN) to capture the long-term dependencies from EMR data. Our model is built upon Diffrentiable Neural Computer (DNC) [6], a recent powerful and fully differentiable MANN. A DNC is an expressive recurrent neural network consisting of a controller augmented with a memory module. At each time step, the controller reads an input, updates the memory, and generates an output. DNC has demonstrated its efficacy in various difficult tasks that require very long chains of computation such as graph prediction and question-answering suggesting its power of solving sequence prediction problems. Despite its potentials, DNC has yet to be applied to healthcare, especially in clinical treatment sequence prediction.

We adapt the DNC to the task of treatment recommendation with two key modifications. We formulate the treatment recommendations as a sequence-to-sequence prediction problem, where the entire medical history sequence stored in EMR is used to produce a sequence of treatment options. The output sequence allows modeling dependencies between current treatments, and between treatment and the distant history. We modify the DNC by using two controllers to handle dual processes: history encoding and treatment recommendations. Each controller will employ different "remembering" strategies for each process helping improvement in prediction and increasing the learning speed. In the second modification, we apply a write-protected policy for the decoding controller, that is, memory is read-only in the decoding phase.

In summary, our main contributions are: (i) handling long-term dependencies in treatment recommendations by solving the sequence prediction problem, (ii) proposing a novel memory-augmented architecture that uses dual controllers and write-protected mechanism (DCw-MANN) to suit sequence-to-sequence task, (iii) empirically evaluating our model on a real-world EMR dataset (MIMIC-III) and showing that our method outperforms existing methods in treatment recommendations. The significance of DCw-MANN lies in its versatility as our model can be applied to other sequential domains with similar data characteristics.

---

[1] https://mimic.physionet.org/mimictables/

## 2 Methods

### 2.1 Problem Formulation

In EMR data, a hospital visit is documented as one admission record consisting of diagnosis and treatment codes for the admission. Diagnoses are coded using WHO's ICD (International Classification of Diseases) coding schemes. For example, in the ICD10 scheme, E10 encodes Type 1 diabetes mellitus, E11 encodes Type 2 diabetes mellitus and F32 indicates depressive episode. The treatment can be procedure or drug. The procedures are typically coded in CPT (Current Procedural Terminology) or ICHI (International Classification of Health Interventions) schemes. The drugs are often coded in ATC (Anatomical Therapeutic Chemical) or NDC (National Drug Code). Once diagnoses are confirmed, we want to predict the output sequence (treatment codes of the current visit) given the input sequence (all diagnoses followed by treatments from the first visit to the previous visit plus the diagnoses of the current visit). More formally, we denote all the unique medical codes (diagnosis, procedures and drugs) from the EMR data as $c_1, c_2, ..c_{|C|} \in C$, where $|C|$ is the number of unique medical code. A patient's $n$-th admission's input is represented by a sequence of codes:

$$\left[c_{d_1}^1, c_{d_2}^1, ..., \permil, c_{p_1}^1, c_{p_2}^1..., \varnothing, ..., c_{d_1}^{n-1}, c_{d_2}^{n-1}, ..., \permil, c_{p_1}^{n-1}, c_{p_2}^{n-1}, ...., \varnothing, c_{d_1}^n, c_{d_2}^n, ..., \permil\right] \quad (1)$$

Here, $c_{d_j}^k$ and $c_{p_j}^k$ are the $j$-th diagnosis and treatment code of the $k$-th admission, respectively. $\permil$, $\varnothing$ are special characters that informs the model about the change from diagnosis to treatment codes in an admission and the end of an admission, respectively. This reflects the natural structure of a medical history, which is a sequence of clinic visits, each of which typically includes a subsequence of diagnoses, and a subset of treatments. A diagnosis subsequence usually started with the primary condition, followed by secondary conditions. In a subset of treatments, the order is not strictly enforced, but it may reflect the coding practice. The output of the patient's $n$-th admission is : $\left[c_{p_1}^n, c_{p_2}^n, ..., c_{p_{L_{out}}}^n, \varnothing\right]$, in which $L_{out}$ is the length of the treatment sequence we want to predict and $\varnothing$ is used to inform the model to stop predicting. Finally, each code is represented by one-hot vector $v_c \in [0, 1]^{\|C\|}$, where $v_c = [0, ..., 0, 1, 0.., 0]$ ($v_c[i] = 1$ if and only if $v_c$ represents $c_i$). Unlike set encoding of each admission, representing the data this way preserves the admission's internal order information, which allows sequence-based methods to demonstrate their power of capturing sequential events.

### 2.2 DNC Overview

In this subsection, we briefly review DNC [6]. A DNC consists of a controller, which accesses and modifies an external memory module using a number of read and write heads. In DNC, the memory module is more powerful and can "remember" a longer sequence than recurrent neural nets such as LSTM [8] or other MANNs such as NTM [5]. Given some input $x_t$, and a set of $R$ read values from memory $r_{t-1} = \left[r_{t-1}^1, r_{t-1}^2, ..., r_{t-1}^R\right]$, the controller produces the output: $o_t \in \mathbb{R}^{|C_p|}$, where $|C_p|$ is the number of possible output and the key $k_t \in \mathbb{R}^D$, where $D$ is the word size in memory. This key will be used for locating the read and write memory slots of a memory matrix $M_t$. The addressing mechanism is mostly based on cosine similarity:

$$D\left(M_t(i), k_t\right) = \frac{k_t \cdot M_t(i)}{||k_t|| \cdot ||M_t(i)||} \quad (2)$$

which is used to produce a content-based read-weight and write-weight vector $w_t^{cr}, w_t^{cw} \in \mathbb{R}^N$ whose elements are computed according to a softmax over memory's locations. $N$ is the number of memory locations. In addition to content-based addressing, DNC supports dynamic memory allocation and temporal memory linkage for computing the final write-weight and read-weight.

**Dynamic memory allocation & write weightings:** DNC maintains a memory usage vector $u_t \in [0,1]^N$ to define the allocation write-weight:

$$a_t\left[\Phi_t\left[j\right]\right] = \left(1 - u_t\left[\Phi_t\left[j\right]\right]\right) \prod_{i=1}^{j-1} u_t\left[\Phi_t\left[i\right]\right] \quad (3)$$

in which, $\Phi_t$ contains elements from $u_t$ sorted by ascending order from least to most used. Given the write gate $g_t^w$ and allocation gate $g_t^a$, the final write-weight then can be computed by interpolating between the content-based write-weight and the allocation write-weight:

$$w_t^w = g_t^w \left[g_t^a a_t + \left(1 - g_t^a\right) w_t^{cw}\right] \quad (4)$$

**Temporal memory linkage & read weightings:** DNC uses a temporal link matrix $L_t \in [0,1]^{N \times N}$ to keep track of consecutively modified memory locations, and $L_t[i,j]$ represents the degree to which location $i$ was the location written to after location $j$. Each time a memory location is modified, the link matrix is updated to remove old links to and from that location, and add new links from the last-written location. Given the link matrix, the final read-weight is given as follow:

$$w_t^{rk} = \pi_t^k\left[1\right] L_t^\top w_{t-1}^{rk} + \pi_t^k\left[2\right] w_t^{crk} + \pi_t^k\left[3\right] L_t w_{t-1}^{rk} \quad (5)$$

The read mode weight $\pi_t^k$ is used to balance between the content-based read-weight and the forward $L_t w_{t-1}^{rk}$ and backward $L_t^\top w_{t-1}^{rk}$ of the previous read. Then, the $k$-th read value $r_t^k$ is retrieved using the final read-weight vector:

$$r_t^k = \sum_i^N w_t^{rk}(i) M_t(i) \quad (6)$$

### 2.3 Proposed Model

We now present our main contribution to solve the task of treatment recommendations – a deep neural architecture called Dual Controller Write-Protected Memory

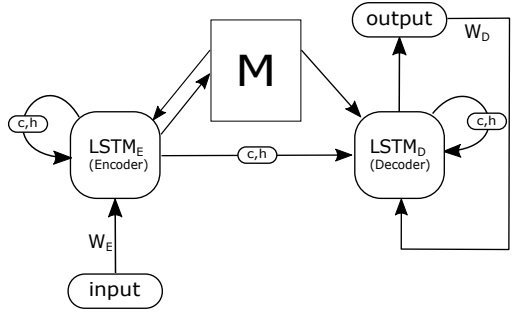

**Fig. 1.** Dual Controller Write-Protected Memory Augmented Neural Network. $LSTM_E$ is the encoding controller implemented as a LSTM. $LSTM_D$ is the decoding controller.

Augmented Neural Network (DCw-MANN)(see Fig. 1). Our DCw-MANN introduces two simple but crucial modifications to the original DNC: (i) using two controllers to handle dual processes of encoding and decoding, respectively; and (ii) applying a write-protected policy in the decoding phase.

In the encoding phase, after going through embedding layer $W_E$, the input sequence is fed to the first controller (encoder) $LSTM_E$. At each time step, the controller reads from and writes to the memory information necessary for the later decoding process. In the decoding phase, the states of the first controller is passed to the second controller (decoder) $LSTM_D$. The use of two controllers instead of one is important in our setting because it is harder for a single controller to learn many strategies at the same time. Using two controllers will make the learning easier and more focused. Also different from the encoder, the decoder can make use of its previous prediction (after embedding layer $W_D$) as the input together with the read values from the memory. Another important feature of DCw-MANN is its write-protected mechanism in the decoding phase. This has an advantage over the writing strategy used in the original DNC since at decoding step, there is no new input that is fed into the system. Of course, there remains dependencies among codes in the output sequence. However, as long as the dependencies among output codes are not too long, they can be well-captured by the cell memory $c_t$ inside the decoder's LSTM. Therefore, the decoder in our design is prohibited from writing to the memory. To be specific, at time step $t+1$ we have the hidden state and cell memory of the controllers calculated as:

$$h_{t+1}, c_{t+1} = \begin{cases} LSTM_E\left([W_E v_{d_t}, r_t], h_t, c_t\right); & t \leq L_{in} \\ LSTM_D\left([W_D v_{p_t}, r_t], h_t, c_t\right); & t > L_{in} \end{cases} \quad (7)$$

where $v_{d_t}$ is the one-hot vector representing the input sequence's code at time $t \leq L_{in}$ and $v_{p_t}$ is the predicted one-hot vector output of the decoder at time $t > L_{in}$, defined as $v_{p_t} = onehot\left(o_t\right)$, i.e.,:

$$v_{p_t}[i] = \begin{cases} 1 & ; i = \underset{1 \leqslant j \leqslant |C_p|}{argmax}(o_t[j]) \\ 0 & ; otherwise \end{cases}. \quad (8)$$

We propose a new memory update rule to enable the write-protected mechanism:

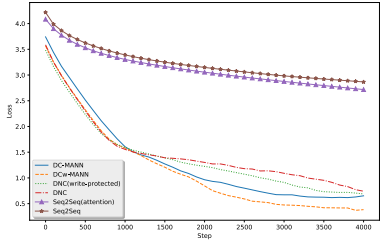 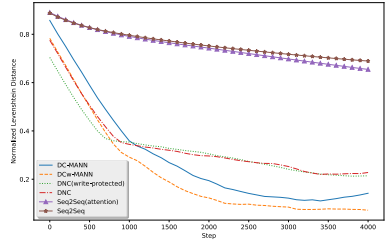

**Fig. 2.** Training Loss of Odd-Even Task    **Fig. 3.** Training NLD of Odd-Even Task

$$M_t = \begin{cases} M_{t-1} \circ \left(E - w_t^w e_t^\top\right) + w_t^w v_t^\top; & t \leq L_{in} \\ M_{t-1}; & t > L_{in} \end{cases} \quad (9)$$

where $E$ is an $N \times D$ matrix of ones, $w_t^w \in [0,1]^N$ is the write-weight, $e_t \in [0,1]^D$ is an erase vector, $v_t \in \mathbb{R}^D$ is a write vector, $\circ$ is point-wise multiplication, and $L_{in}$ is the length of input sequence.

## 3  Results

In this section, we perform experiments both on real-world data and synthetic tasks. The purpose of the synthetic task is to study the incremental impact of modifications we propose.

### 3.1  Synthetic Task: Odd-Even Sequence Prediction

In this task, the input is sequence of random odd numbers chosen without replacement from the set $S_o = \{1, 3, 5, ..., 49\}$ and the output is sequence of even numbers from the set $S_e = \{2, 4, 6, ..98\}$. The $n$-th number $y_n$ in the output sequence is computed as:

$y_n = \begin{cases} 2x_n & n \leq \lfloor \frac{L}{2} \rfloor \\ y_{n-1} + 2 & n > \lfloor \frac{L}{2} \rfloor \end{cases}$. $x_n$ is the $n$-th number in the input sequence and $L$ is the length of both input and output sequence chosen randomly from the range $[1, 20]$. The formula is designed to reflect healthcare situations where treatment options depend both on diagnoses in the input sequence and other treatments in the same output sequence. Here is an example of an input-output sequence pair with $L = 7$: $input := [11, 7, 25, 39, 31, 1, 13]$ and $output := [22, 14, 50, 52, 54, 56, 58]$. We want to predict the even numbers in the output sequence given odd numbers in the input sequence, hence we name it odd-even prediction task. In this task, the model has to "remember" the first half of the input sequence to compute the first half of the output sequence, then it should switch from using input to using previous output at the middle of the output sequence to predict the second half.

| Model | NLD |
|---|---|
| Seq2Seq | 0.679 |
| Seq2Seq with attention | 0.637 |
| DNC | 0.267 |
| DNC (write-protected) | 0.250 |
| DC-MANN | 0.161 |
| DCw-MANN | **0.082** |

**Table 1.** Test Results on Odd-Even Task (lower is better)

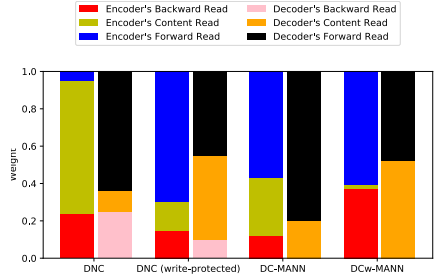

**Fig. 4.** Read Modes of MANNs on Odd-Even Task

**Evaluations:** Our baselines are Seq2Seq [20], its attention version [1] and the original DNC [6]. Since we want to analyze the impact of new modifications, in this task, we explore two other models: DNC with write-protected mechanism in the decoding phase and dual controller MANN without write-protected mechanism (DC-MANN). We use the Levenshtein distance (edit distance) to measure the model's performance. To account for variable sequence lengths, we normalize this distance over the length of the longer sequence (between 2 sequences). The predicted sequence is good if its Normalized Levenshtein Distance (NLD) to the target sequence is small.

**Implementation Details:** For all experiments, deep learning models are implemented in Tensorflow 1.3.0. Optimizer is Adam [10] with learning rate of 0.001 and other default parameters. The hidden dimensions for LSTM and the embedding sizes for all models are set to 256 and 64, respectively. Memory's parameters including number of memory slots and the size of each slot are set to 128 and 128 , respectively.

**Results:** After training with 4000 input-output pair of sequences, the models will be tested for the next 1000 pairs. The learning curves of the models are plotted in Figs. 2 and 3. The average NLD of the predictions is summarized in Table 1. As is clearly shown, the proposed model outperforms other methods. Seq2Seq-based methods fail to capture the data pattern and underperform other methods. The introduction of two controllers helps boost the performance of DNC significantly. Additional DNC-variant with write-protected also performs better than the original one, which suggests the benefit of decoding without writing.

Fig. 4 plots read mode weights for three reading strategies employed in encoding and decoding phases. We can observe the differences in the way the models prefer reading strategies. The biggest failure of DNC is to keep using backward read in the decoding process. This is redundant because in this problem, it is the forward of the previous read location (if the memory location that corresponds to $x_{n-1}$ is the previous read, then its forward is the memory location that corresponds to $x_n$) that defines the current output ($y_n$). On the other hand, dual controllers with write-protected mechanism seems help the model avoid bad strategies and focus more on learning reasonable strategies. For example, using dual controllers tends to lessen the usage of content-based read in the encoding phase. This strategy is reasonable in this example since the input at each time step is not repeated. Write-protected policy helps balance the forward and content-based read in the decoding phase, which may reflect the output pattern – half-dependent on the input and half-dependent on the previous output.

| MIMIC-III Dataset (# of visit >1) | Procedure as output | Drug as output |
|---|---|---|
| # of patients | 6,314 | 5,620 |
| # of admissions | 16,317 | 14,656 |
| # of unique diagnosis codes | 4,669 | 4,563 |
| # of unique treatment codes | 1,439 | 2,446 |
| Average # of diagnosis sequence length | 13.3 | 13.8 |
| Max # of diagnosis sequence length | 39 | 39 |
| Average # of treatment sequence length | 4.7 | 11.4 |
| Max # of treatment sequence length | 40 | 186 |
| Average # of visits per patient | 2.5 | 2.6 |
| Max # of visits per patient | 29 | 29 |

**Table 2.** Statistic of MIMIC-III sub-datasets

### 3.2 Treatment Recommendation Tasks

The dataset used for this task is MIMIC-III [9], which is a publicly available dataset consisting of more than 58k EMR admissions from more than 46k patients. An admission history in this dataset can contain hundreds of medical codes, which raises a great challenge in handling long-term dependencies. In MIMIC-III, there are both procedure and drug codes for the treatment process so we consider two separate treatment recommendation tasks: procedure prediction and drug prescription. In practice, if we use all the drug codes in an EMR record, the drug sequence can be very long since, each day in hospital, the doctor can prescribe several types of drugs for the patient. Hence, we only pick the first drug used in a day during the admission as the representative drug for that day. We also follow the previous practice that only focuses on patients who have more than one visit [13,14,15]. The statistics of the two sub-datasets is detailed in Table 2.

**Evaluations:** For comprehensiveness, beside direct competitors, we also compare our methods with classical for healthcare predictions, which are Logistic Regression and Random Forests. Because traditional methods are not designed for sequence predictions, we simply pick the top outputs and ignore their orders. In treatment recommendation tasks, we use precision, which is defined as the number of correct predicted treatment codes (ignoring the order) divided by the number of predict treatment codes. More formally, let $S_p^n$ be the set of ground truth treatments for the $n$-th admission, $S_q^n$ be the set of treatments that the model outputs. Then the precision is: $\frac{1}{N} \sum_{n=1}^{N} \frac{|S_p^n \cap S_q^n|}{|S_q^n|}$, where $N$ is total number of test patients. To measure how closely the generated treatment compares against the real treatment, we use Mean Jaccard Coefficient[2], which is defined as the size of the intersection divided by the size of the union of ground truth treatment set and predicted treatment set: $\frac{1}{N} \sum_{n=1}^{N} \frac{|S_p^n \cap S_q^n|}{|S_p^n \cup S_q^n|}$.

| Model | Procedure Output | | Drug Output | |
|---|---|---|---|---|
| | Precision | Jaccard | Precision | Jaccard |
| Logistic Regression | 0.256 | 0.185 | 0.412 | 0.311 |
| Random Forest | 0.276 | 0.199 | 0.491 | 0.405 |
| Seq2Seq | 0.263 | 0.196 | 0.220 | 0.138 |
| Seq2Seq with attention | 0.272 | 0.204 | 0.224 | 0.142 |
| DNC | 0.285 | 0.214 | 0.577 | 0.529 |
| DCw-MANN | **0.292** | **0.221** | **0.598** | **0.556** |

**Table 3.** Results on MIMIC-III dataset for procedure prediction and drug prescription (higher is better).

**Implementation Details:** We randomly divide the dataset into the training, validation and testing set in a $0.7 : 0.1 : 0.2$ ratio, where the validation set is used to tune model's hyper-parameters. For the classical Random Forests and Logistic Classifier, the input is bag-of-words. Also, we apply One-vs-Rest strategy [17] to enable these classifiers to handle multi-label output and the hyper-parameters are found by grid-searching.

**Results:** Table 3 reports the prediction results on two tasks (procedure prediction and drug prescription). The performance of the proposed DCw-MANN is higher than that of baselines on the testing data for both tasks, validating the use of dual controllers with write-protected mechanism. Without memory, Seq2Seq methods seem unable to outperform classical methods, possibly because the evaluations are set-based, not sequence-based. In the drug prescription task, there is a huge drop in performance of the Seq2Seq-based approaches. It should be noted that, in drug prescription, the drug codes are given day by day; hence, the average length of output sequence are much longer than the procedure's one. This could be a very challenging task for Seq2Seq. Memory-augmented models, on the other hand, have an external memory to store information, so it can cope with long-term dependencies. Figs. 5 and Fig. 6 show that compared to DNC, DCw-MANN is the faster learner in drug prescription task. This case study demonstrates that a MANN with dual controller and write-protected mechanism can significantly improve the performance of the sequence prediction task in healthcare.

---

[2] The metrics actually are at disadvantage to the proposed sequence-to-sequence model, but we use to make them easy to compare against non-sequential methods.

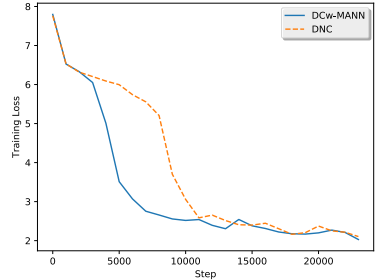 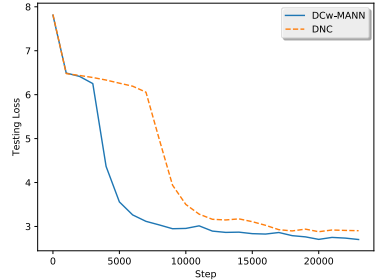

**Fig. 5.** Training Loss of Drug Prescription Task

**Fig. 6.** Testing Loss of Drug Prescription Task

## 4 Related Works

The recent success of deep learning has drawn board interest in building AI systems to improve healthcare. Several studies have used deep learning methods to better categorize diseases and patients: denoising autoencoders, an unsupervised approach, can be used to cluster breast cancer patients [21], and convolutional neural networks (CNNs) can help count mitotic divisions, a feature that is highly correlated with disease outcome in histological images [4]. Another branch of deep learning in healthcare is to solve biological problems such as using deep RNN to predict gene targets of microRNAs [24]. Despite these advances, a number of challenges exist in this area of research, most notably how to make use of other disparate types of data such as electronic medical records (EMRs). Recently, more efforts have been made to utilize EMR data in disease prediction [15], unplanned admission and risk prediction [14] problems. Other works apply LSTMs, both with and without attention to clinical time series for heart failure prediction [3] or diagnoses prediction [11]. Treatment recommendation is also an active research field with recent deep learning works that model EMR codes as sequence such as [2] using sequence of billing codes for medicine suggestions or [23] using set of diagnoses for medicine sequence prediction. Differing from these approaches, our work focuses on modeling both the admission data and the treatment output as two sequences to capture order information from input codes and ensure dependencies among output codes at the same time.

Memory augmented neural networks (MANN) have emerged as a new promising research topic in deep learning. Memory Networks (MemNNs) [22] and Neural Turing Machines (NTMs) [5] are the two classes of MANNs that have been applied to many problems such as meta learning [18] and question answering [19]. In healthcare, there is limited work applying MemNN-based models to handle medical-related problems such as clinical textual QA [7] or diagnosis inference [16]. However, these works have been using clinical documents as input, rather than just using medical codes stored in EMRs. Our work, on the other hand, learns end-to-end from raw medical codes in EMRs by leveraging Differentiable Neural Computer (DNC) [6], the latest improvement over the NTM. In practice, DNC and other NTM variants have been used for various domains such as visual question answering [12], and one-shot learning [18], yet it is the first time DNC is adapted for healthcare tasks.

## 5  Conclusion

We have introduced a dual controller write-protected MANN designed for healthcare treatment recommendations. Under our design, the order dependencies inside each admission and between admissions are preserved, which allows memory-based methods to make use of this sequential information for better performance. Differing from other approaches, our work is one of the first attempts to apply MANN to healthcare domain and promising results on MIMIC-III dataset have shown that modifications such as using two controllers and write-protected mechanism are necessary to make MANN work for real-world problems like treatment prediction. In additions, our method can be generalized to other sequence prediction tasks that require special handling of long-term dependencies. Future work will focus on extending the model to handle multiple healthcare tasks, and developing new capabilities for medical question answering.

## References

1. D. Bahdanau, K. Cho, and Y. Bengio. Neural Machine Translation by Jointly Learning to Align and Translate. *ICLR*, 2015.
2. Jacek M Bajor and Thomas A Lasko. Predicting medications from diagnostic codes with recurrent neural networks. *ICLR*, 2017.
3. Edward Choi, Mohammad Taha Bahadori, Jimeng Sun, Joshua Kulas, Andy Schuetz, and Walter Stewart. RETAIN: An Interpretable Predictive Model for Healthcare using Reverse Time Attention Mechanism. In *Advances in Neural Information Processing Systems*, pages 3504–3512, 2016.
4. Dan C Cireşan, Alessandro Giusti, Luca M Gambardella, and Jürgen Schmidhuber. Mitosis detection in breast cancer histology images with deep neural networks. In *International Conference on Medical Image Computing and Computer-assisted Intervention*, pages 411–418. Springer, 2013.
5. A. Graves, G. Wayne, and I. Danihelka. Neural Turing Machines. *ArXiv e-prints*, October 2014.
6. Alex Graves, Greg Wayne, Malcolm Reynolds, Tim Harley, Ivo Danihelka, Agnieszka Grabska-Barwińska, Sergio Gómez Colmenarejo, Edward Grefenstette, Tiago Ramalho, John Agapiou, et al. Hybrid computing using a neural network with dynamic external memory. *Nature*, 538(7626):471–476, 2016.
7. Sadid A Hasan, Siyuan Zhao, Vivek V Datla, Joey Liu, Kathy Lee, Ashequl Qadir, Aaditya Prakash, and Oladimeji Farri. Clinical question answering using key-value memory networks and knowledge graph. In *TREC*, 2016.
8. Sepp Hochreiter and Jürgen Schmidhuber. Long short-term memory. *Neural computation*, 9(8):1735–1780, 1997.

9. Alistair EW Johnson, Tom J Pollard, Lu Shen, Li-wei H Lehman, Mengling Feng, Mohammad Ghassemi, Benjamin Moody, Peter Szolovits, Leo Anthony Celi, and Roger G Mark. MIMIC-III, a freely accessible critical care database. *Scientific data*, 3, 2016.
10. Diederik Kingma and Jimmy Ba. Adam: A method for stochastic optimization. *arXiv preprint arXiv:1412.6980*, 2014.
11. Z. Lipton, D. Kale, C. Elkan, and R. Wetzel. Learning to Diagnose with LSTM Recurrent Neural Networks. In *International Conference on Learning Representations (ICLR 2016)*, 2016.
12. Chao Ma, Chunhua Shen, Anthony Dick, and Anton van den Hengel. Visual question answering with memory-augmented networks. *arXiv preprint arXiv:1707.04968*, 2017.
13. Fenglong Ma, Radha Chitta, Jing Zhou, Quanzeng You, Tong Sun, and Jing Gao. Dipole: Diagnosis prediction in healthcare via attention-based bidirectional recurrent neural networks. In *Proceedings of the 23rd ACM SIGKDD International Conference on Knowledge Discovery and Data Mining*, pages 1903–1911. ACM, 2017.
14. Phuoc Nguyen, Truyen Tran, Nilmini Wickramasinghe, and Svetha Venkatesh. Deepr: A Convolutional Net for Medical Records. *Journal of Biomedical and Health Informatics*, 21(1), 2017.
15. Trang Pham, Truyen Tran, Dinh Phung, and Svetha Venkatesh. Predicting healthcare trajectories from medical records: A deep learning approach. *Journal of Biomedical Informatics*, 69:218–229, May 2017.
16. Aaditya Prakash, Siyuan Zhao, Sadid A Hasan, Vivek V Datla, Kathy Lee, Ashequl Qadir, Joey Liu, and Oladimeji Farri. Condensed memory networks for clinical diagnostic inferencing. In *AAAI*, pages 3274–3280, 2017.
17. Ryan Rifkin and Aldebaro Klautau. In defense of one-vs-all classification. *Journal of machine learning research*, 5(Jan):101–141, 2004.
18. Adam Santoro, Sergey Bartunov, Matthew Botvinick, Daan Wierstra, and Timothy Lillicrap. Meta-learning with memory-augmented neural networks. In *International conference on machine learning*, pages 1842–1850, 2016.
19. Sainbayar Sukhbaatar, Jason Weston, Rob Fergus, et al. End-to-end memory networks. In *Advances in neural information processing systems*, pages 2440–2448, 2015.
20. Ilya Sutskever, Oriol Vinyals, and Quoc V. Le. Sequence to sequence learning with neural networks. *International Conference on Machine Learning*, 2014.
21. Jie Tan, Matthew Ung, Chao Cheng, and Casey S Greene. Unsupervised feature construction and knowledge extraction from genome-wide assays of breast cancer with denoising autoencoders. In *Pacific Symposium on Biocomputing Co-Chairs*, pages 132–143. World Scientific, 2014.
22. Jason Weston, Sumit Chopra, and Antoine Bordes. Memory networks. *arXiv preprint arXiv:1410.3916*, 2014.
23. Yutao Zhang, Robert Chen, Jie Tang, Walter F Stewart, and Jimeng Sun. Leap: Learning to prescribe effective and safe treatment combinations for multimorbidity. In *Proceedings of the 23rd ACM SIGKDD International Conference on Knowledge Discovery and Data Mining*, pages 1315–1324. ACM, 2017.
24. Jozef Zurada. End effector target position learning using feedforward with error backpropagation and recurrent neural networks. In *Neural Networks, 1994. IEEE World Congress on Computational Intelligence., 1994 IEEE International Conference on*, volume 4, pages 2633–2638. IEEE, 1994.